\documentclass[runningheads]{llncs}

\usepackage{eccv}

\usepackage{eccvabbrv}

\usepackage{graphicx}
\usepackage{booktabs}
\usepackage{microtype}
\usepackage{xcolor}
\usepackage{multirow}

\usepackage[accsupp]{axessibility}  

\usepackage{hyperref}

\usepackage{orcidlink}

\newcommand{\ours}{NeuroGuard (AGS+CRK+FBE)}

\begin{document}

\title{NeuroGuard: Neural Gradient Update Aware of Representation Damage}
\titlerunning{NeuroGuard: Gradient Update Aware of Representation Damage}

\author{Taigo Sakai\inst{1}\orcidlink{0009-0000-5130-1522} \and
Kazuhiro Hotta\inst{1}\orcidlink{0000-0002-5675-8713}}

\authorrunning{T.Sakai et al.}

\institute{Meijo University, 1-501 Shiogamaguchi, Tempaku-ku, Nagoya 468-8502, Japan \email{200442066@ccalumni.meijo-u.ac.jp} \email{kazuhotta@meijo-u.ac.jp}\\}

\maketitle

\begin{abstract}
Long-tailed class-incremental learning (LT-CIL) must learn new classes from imbalanced streams while retaining old classes. Existing methods mainly change replay, classifiers, or losses. We study a different factor, namely how strongly the feature representation should be updated at each task boundary.
We propose NeuroGuard, an update-control method added to DGR, a replay-based LT-CIL baseline, without adding learnable parameters. NeuroGuard preserves DGR's replay memory, classifier, and set of loss terms. Adaptive Gradient Scaling (AGS) converts teacher uncertainty into one task-wise gradient scale.
Confidence-Ranked Knowledge Distillation Reweighting (CRK) gives larger knowledge-distillation weights to replay samples that the teacher predicts less decisively.
Fragility-Blended Entropy Gate (FBE) adds old-memory leakage to the scale decision.
Across five LT-CIL settings, NeuroGuard improves over DGR in every setting. In the four main benchmark comparisons, it achieves the best task-agnostic accuracy among the compared methods. The gains extend to both old- and new-class accuracy, while medium-frequency accuracy improves consistently across all five settings. Controlled comparisons show that the gain does not come from generic gradient suppression: AGS outperforms a matched fixed-scale control in all five settings, demonstrating that boundary-specific scaling is more effective than applying the same average scale throughout learning.

\end{abstract}

\section{Introduction}
\label{secintro}

Natural data streams exhibit two forms of non-stationarity.
First, class frequencies are highly imbalanced. A small number of classes appear frequently while many others remain rare.
Second, new classes arrive sequentially over time rather than being observed simultaneously.
Class-incremental learning (CIL) addresses the sequential aspect of this problem by learning new classes while retaining previously acquired knowledge~\cite{rebuffi2017icarl,hou2019learning}.
Long-tailed class-incremental learning (LT-CIL) additionally considers severe class-frequency imbalance~\cite{liu2022longtailed}.
These two factors interact during optimization.
Majority classes dominate gradient statistics, minority classes receive weaker learning signals, and updates introduced by later tasks can overwrite representations formed for earlier tasks~\cite{liu2022longtailed,kim2020imbalanced}.
As a result, LT-CIL requires balancing adaptation to new information against preservation of existing knowledge.

Biological learning systems offer a useful perspective on this trade-off.
In the brain, a small brainstem nucleus called the locus coeruleus responds to unexpected inputs by releasing a chemical signal that transiently modulates neural responsiveness across the cortex~\cite{astonjonescohen2005,yudayan2005}.
In short, one simple ``surprise'' signal turns the size of neural updates up or down, without changing the network itself.
We use the same idea for LT-CIL. A single number measured at each task boundary controls how strongly the representation is updated, while the classifier, replay memory, and set of loss terms remain unchanged. We therefore focus on two complementary questions. The first is how much the representation should change at a task boundary. The second is which replayed memories deserve stronger protection during distillation.

To answer these two questions, we do not introduce a new network structure or an additional loss term. We add a lightweight update-control rule to an existing replay-based method.
As shown in Fig.~\ref{figoverview}, we build NeuroGuard on DGR~\cite{dgr}, a replay-based LT-CIL baseline that reweights classifier gradients by class frequency, while preserving its replay memory, classifier, and set of loss terms.
NeuroGuard has three components, none of which adds learnable parameters. AGS decides how strongly the representation is updated at each task boundary. CRK decides which replayed memories receive stronger distillation. FBE connects these two decisions by measuring old-memory leakage and incorporating it into the scale decision.
The components are added progressively.

On ImageNet-100-LT, the full model reaches 34.0\% task-agnostic accuracy, improving over the DGR baseline by $0.99$ \% while outperforming DER, FOSTER, and COIL (Table~\ref{tabimagenet}).
On CIFAR-100-LT \texttt{10base10}, it achieves the highest reported task-agnostic accuracy among the compared methods, exceeding DER by $2.35$ \%.
On the harder \texttt{50base11} split, our method improves over DGR by $2.35$ \% and exceeds DER by $0.32$ \%, achieving the strongest result in this setting without model expansion.
On Food-101-LT, our method also gives the best task-agnostic accuracy, improving over DGR by $1.09$ \%, with the largest subgroup gains on new- and minor-class accuracy.
To better understand these outcomes, we additionally compare AGS with a fixed-scale control.
The fixed-scale control uses $g=0.864$, the arithmetic mean obtained by pooling all task-wise gradient scales produced by AGS across the five evaluated settings. It remains below AGS in all five settings and falls below DGR in three. Because this control removes only boundary-dependent scale selection while retaining the same gradient-scaling mechanism, the advantage observed in all five settings directly supports task-boundary-specific adaptation. A single constant cannot reproduce the benefit across changing class streams.

The contributions of this paper are as follows.
\begin{itemize}
\item[(1)]{We introduce Adaptive Gradient Scaling (AGS), a task-boundary update rule that controls representation updates using teacher uncertainty without adding learnable parameters.}
\item[(2)]{We propose CRK, which prioritizes fragile replayed memories while keeping the mean distillation weight equal to one, and FBE, which augments the boundary signal with old-memory leakage. Neither component changes the replay memory or classifier structure.}
\item[(3)]{We provide a progressive analysis of AGS, CRK, and FBE, together with a fixed-scale control that uses the mean task-wise AGS scale across all evaluated settings to separate boundary-dependent scale selection from generic gradient suppression.}
\item[(4)]{We evaluate the method on five LT-CIL settings and show that AGS outperforms this fixed-scale control in every setting. This consistent result supports adapting update strength to each task boundary rather than using one universal constant.}
\end{itemize}

\section{Related Work}
\label{secrelated}

\paragraph{Long-tailed class-incremental learning.}

CIL methods learn new classes without retaining the full past-task training sets~\cite{rebuffi2017icarl,hou2019learning,masana2023survey}.
We follow the standard class-incremental scenario~\cite{vandeven2022three,mundt2021essentials,delange2021isCILenough}.
In LT-CIL, class frequencies are imbalanced, so majority classes dominate gradient statistics and minority classes are harder to learn~\cite{liu2022longtailed,kim2020imbalanced}.
Recent methods rebalance replay~\cite{liu2022longtailed,er2019}, adjust logits~\cite{menon2021longtail}, reweight distillation~\cite{kim2020imbalanced,kang2021balancedkd}, or modify batch normalization~\cite{tbbn2022}.
Long-tailed recognition beyond CIL has also been studied with mutual-information and expert-based objectives~\cite{suh2023ltmi,mdcs2023}.
DGR~\cite{dgr}, which we use as the base method, reweights classifier gradients by class for imbalanced CIL.
Expansion-based methods such as DER and FOSTER increase model capacity, and COIL is a strong co-transport baseline~\cite{yan2021der,wang2022foster,coil2021}.
Recent work also revisits CIL evaluation setups~\cite{futureproof2024,opencil2024}.
NeuroGuard differs from these studies by what it leaves unchanged. It preserves the DGR replay rule, classifier, and set of loss terms. It controls only the strength of representation updates and the relative knowledge-distillation weight of replayed old samples. This isolates the effect of update control from extra capacity and additional loss terms.

\paragraph{Stability and plasticity.}

Continual learning must balance new-class learning and old-class retention.
Regularization methods such as EWC~\cite{kirkpatrick2017overcoming} protect important parameters from large updates.
Metaplasticity studies motivate the idea that plasticity should change with prior learning and current uncertainty~\cite{abraham1996metaplasticity,zenke2017continual,laborieux2021synaptic}.
Our proposed AGS follows this motivation. It tests whether one task-wise gradient scale at the final convolutional stage can regulate representation updates.
Our proposed FBE extends this scale decision with a memory-side boundary signal. This signal measures how much probability mass from old exemplars is assigned to newly added class outputs.

\paragraph{Gradient modulation and distillation reweighting.}

Gradient projection methods~\cite{saha2021gradient} and Gradient Episodic Memory~\cite{lopezpaz2017gem} explicitly constrain parameter updates to reduce interference between tasks.
Our proposed AGS has a narrower target. It does not project gradients at every optimization step. It applies one task-wise gradient scale to the final convolutional stage after each task boundary. This keeps scale-rule comparisons simple.
Sample-wise distillation~\cite{primeaware2020}, subspace distillation~\cite{subspaceKD2023}, and related knowledge-distillation studies~\cite{clkdsurvey2023} reweight what to preserve.
Our proposed CRK also reweights preservation, but it keeps the mean distillation weight equal to one. It redistributes weight inside each replay batch according to the teacher-margin rank of each old sample.
Margin-based replay metrics appear in~\cite{pgd_crs2021}.
In short, AGS controls update size. CRK controls which replay samples receive stronger distillation. FBE supplies an old-memory fragility signal for the scale decision. None of these three proposed components adds learnable parameters or changes the replay memory.

\section{Proposed Method}
\label{secmethod}

Building on DGR~\cite{dgr}, we introduce NeuroGuard for LT-CIL.
NeuroGuard has three components, none of which adds learnable parameters. Adaptive Gradient Scaling (AGS) sets one task-wise gradient scale for representation updates. Confidence-Ranked Knowledge Distillation Reweighting (CRK) gives larger knowledge-distillation weights to fragile replay samples. Fragility-Blended Entropy Gate (FBE) adds old-memory leakage to the boundary signal used by AGS.
Conventional DGR is a replay-based LT-CIL baseline that reweights classifier gradients by class while using standard distillation and a fixed exemplar buffer. We preserve its replay memory, classifier, and set of loss terms.
None of our additions introduces learnable parameters, expands the network, or changes the replay memory.

\begin{figure}[t]
    \centering
    \includegraphics[width=0.90\linewidth]{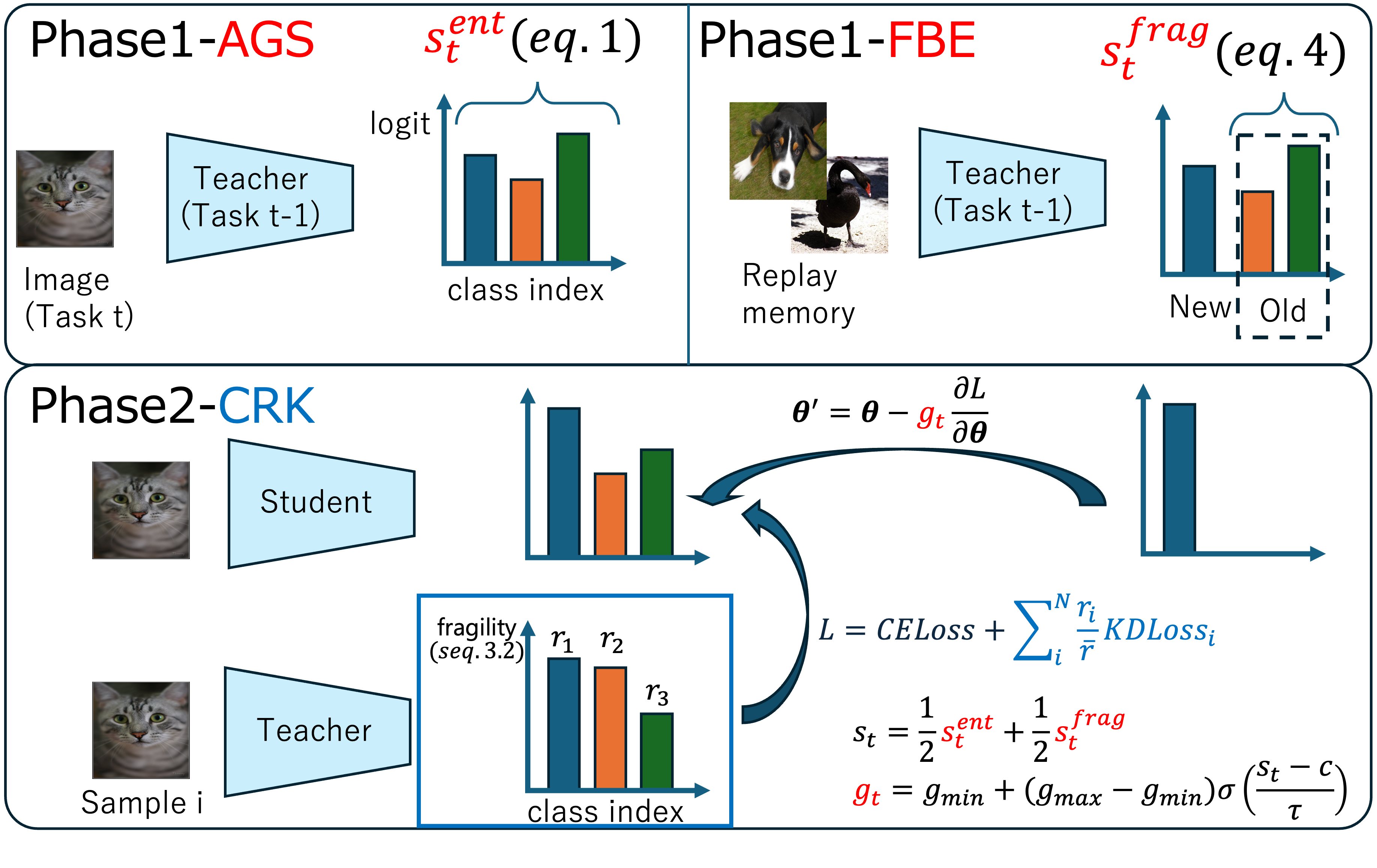}
    \caption{\textbf{Overview of the proposed method.}
    \textbf{First phase.} Before learning task~$t$, the teacher model from task~$(t-1)$ estimates the task boundary. The proposed AGS computes an entropy-based boundary score $s_t^{\mathrm{ent}}$ from current-task samples using Eq.~\eqref{eqentropy}. The proposed FBE then refines this estimate with the fragility score $s_t^{\mathrm{frag}}$ from replay samples in Eq.~\eqref{eqfrag}. The final boundary score is transformed into the task-wise gradient scale $g_t$.
    \textbf{Second phase.} During training, the proposed CRK assigns a confidence-aware weight $w_i$ to each replay sample based on the teacher prediction, and performs weighted knowledge distillation together with task-wise gradient scaling.}
        \label{figoverview}
\end{figure}

\subsection{Adaptive Gradient Scaling (AGS)}

The proposed AGS realizes the neuromodulation-inspired idea from Sec.~\ref{secintro}. A single ``surprise'' score is measured once at each task boundary and sets how strongly the representation is updated.

\paragraph{Teacher uncertainty at a task boundary.}

At the boundary between tasks $t-1$ and $t$, after training on task $t-1$ has finished and before the first update for task $t$, we evaluate the consolidated teacher on a mini-batch of incoming samples. The consolidated teacher is the frozen DGR model trained through task $t-1$ and used for knowledge distillation.
Let $K_{t-1}$ denote the number of classes learned before task $t$.
For each sample, we compute the normalized predictive entropy

\begin{equation}
s_t^{\mathrm{ent}}
=
\frac1B
\sum_{b=1}^{B}
\frac{
-\sum_{k=1}^{K_{t-1}}
p_{\mathrm{old}}(k\mid x_b)
\log p_{\mathrm{old}}(k\mid x_b)
}
{\log K_{t-1}},
\label{eqentropy}
\end{equation}
where
$p_{\mathrm{old}}(k\mid x)$
is the softmax probability predicted by the consolidated teacher.
We divide by $\log K_{t-1}$, the maximum possible entropy over $K_{t-1}$ classes, so that the score always lies in $[0,1]$. Without this normalization, the raw entropy would grow simply because later tasks have more classes, and the boundary scores of early and late tasks would no longer be comparable.
Low entropy indicates that the teacher is confident on the incoming task,
whereas high entropy indicates larger uncertainty.

\paragraph{Task-wise gradient scale.}

The boundary signal is converted into a scalar gradient scale
\begin{equation}
g_t
=
g_{\min}
+
(g_{\max}-g_{\min})
\sigma\left(\frac{s_t-c}{\tau}\right),
\label{eqscale}
\end{equation}
where $\sigma(\cdot)$ is a sigmoid function, $c$ is the center, $\tau$ controls the transition slope, and $g_{\min}$ and $g_{\max}$ define the allowable scale range. During task $t$, only gradients of the final convolutional stage are multiplied by $g_t$.
Here $s_t$ denotes the boundary signal fed into the scale. In the proposed AGS, this signal is the entropy score, so $s_t=s_t^{\mathrm{ent}}$. The proposed FBE later replaces $s_t$ with a blended signal. We therefore keep the separate symbol $s_t$ instead of writing $s_t^{\mathrm{ent}}$ throughout.

As indicated by the curved arrow labeled $g_t$ in the lower-right part of Fig.~\ref{figoverview}, the scale is applied to the backward update during task $t$. For each mini-batch, we compute the ordinary gradients from the DGR loss and then multiply only the gradients of the final convolutional stage by $g_t$ before the optimizer step. Other layers keep their original gradients. Thus, $g_t<1$ suppresses the representation update for the current task, while $g_t\approx1$ gives almost the same update as DGR. Under AGS, the forward computation, loss function, replay memory, and optimizer settings are unchanged.

\subsection{Confidence-Ranked Knowledge Distillation Reweighting (CRK)}

Standard knowledge distillation assigns the same weight to every replayed old exemplar. The proposed CRK changes only this relative weighting. It allocates more distillation weight to replayed samples that the teacher finds less reliable.

For each replayed sample whose label belongs to a previous class, $y_i\in\{1,\ldots,K_{t-1}\}$, the consolidated teacher predicts class probabilities. We define the confidence margin
\[
m_i
=
p_1-p_2,
\]
where $p_1$ and $p_2$ are the largest and second-largest teacher probabilities.
A possible alternative is to define fragility by the probability of the ground-truth class. We do not use this definition because CRK reweights knowledge distillation, not supervised classification. Distillation preserves the teacher's output distribution, so we measure how stable that distribution is. The top-1 versus top-2 margin provides this signal. A small margin means that the teacher is uncertain between two classes, and such a relation can be easily changed during student training. We therefore assign larger distillation weight to replay samples with smaller margins. Fragility is
\[
f_i=1-m_i.
\]

Replayed old samples are ranked by fragility, with rank~1 assigned to the least fragile sample and the largest rank assigned to the most fragile sample. If multiple samples share the same margin, their average rank is used. The knowledge-distillation weight is
\begin{equation}
w_i
=
\frac{r_i}{\bar r},
\label{eqcrk}
\end{equation}
where $r_i$ is the rank and $\bar r$ is the average rank over replayed old samples in the current batch.
Therefore, the mean knowledge-distillation weight is one, while more fragile memories receive larger weights.
Sec.~\ref{sec:ablation} also compares a margin-proportional alternative.

\subsection{Fragility-Blended Entropy Gate (FBE)}

Teacher entropy reflects uncertainty about incoming classes, but it does not measure how the expanded classifier redistributes old-sample probability mass toward new outputs. The proposed FBE adds this memory-side boundary signal.
We measure this score before learning task~$t$. At this point, the classifier head has been expanded with the output units for task~$t$, but the model has not taken any gradient step on task~$t$. Therefore, the backbone and old-class weights are still inherited from task~$t-1$.
The expanded model can place non-negligible probability mass on the newly introduced class logits when evaluating old exemplars.
We define the normalized leakage score
\begin{equation}
s_t^{\mathrm{frag}}
=
\frac1{|M_{\mathrm{old}}|}
\sum_{x\in M_{\mathrm{old}}}
\sum_{k=K_{t-1}+1}^{K_t}
p(k\mid x),
\label{eqfrag}
\end{equation}
where $M_{\mathrm{old}}$ denotes the set of replayed old exemplars and $p(k\mid x)$ is the softmax probability of the expanded model evaluated before training on task $t$.

The boundary signal becomes
\begin{equation}
s_t
=
\frac12
s_t^{\mathrm{ent}}
+
\frac12
s_t^{\mathrm{frag}}.
\label{eqfbe}
\end{equation}

Although the two terms come from different quantities, both lie in $[0,1]$ and quantify mismatch at the task boundary. A larger $s_t^{\mathrm{ent}}$ indicates that incoming samples are less compatible with the consolidated teacher. A larger $s_t^{\mathrm{frag}}$ indicates that the expanded classifier assigns more old-sample probability mass to new outputs. We average the two scores without tuning a mixing weight.

Consequently, the update scale stays near $g_{\min}$ when both boundary signals are low, and it increases as either source of boundary mismatch becomes stronger.

\subsection{Fixed-Scale Control}
\label{sec:fixedscale}

To separate adaptive scale selection from the scale value itself, we additionally evaluate a fixed-scale control.
This baseline removes the adaptive part of AGS. It keeps the same training procedure and applies one constant scale $g$ to the gradients of the final convolutional stage for every task. We set $g=0.864$, the arithmetic mean obtained by pooling all task-wise gradient scales from the completed AGS runs across the five evaluated settings. This gives the fixed-scale control direct access to the average suppression selected by the adaptive method and tests whether matching that average is sufficient. Its lower accuracy in all five settings demonstrates that the mean scale alone cannot reproduce the benefit of boundary-dependent scaling.

\section{Experiments}
\label{secexperiment}

\subsection{Experimental Settings}

We evaluate five LT-CIL settings. Three settings use CIFAR-100-LT~\cite{cifar100lt,liu2022code} with ResNet-32. They are \texttt{10base10} (10 tasks of 10 classes), \texttt{50base11} (50 base classes followed by ten increments of five classes), and \texttt{50base6} (50 base classes followed by five increments of ten classes). We use \texttt{50base6} as a case study for the progressive ablation. We adopt the LT-CIL task splits from the official implementation~\cite{liu2022code}.
Two settings use ResNet-18. ImageNet-100-LT uses \texttt{10base10}, comprising 10 tasks of 10 classes. Food-101-LT uses \texttt{11base10}, comprising 11 base classes followed by nine increments of 10 classes~\cite{ltfood2023}.
All datasets use an exponential profile with imbalance ratio 100 (factor $0.01$).
Replay is fixed at five exemplars per class, aligning with memory-constrained continual-learning scenarios~\cite{clapps2023,er2019}.

CIFAR runs use 160 epochs (milestones 80, 120). ImageNet and Food-101 runs use 40 epochs (milestones 25, 35). Weight decay is $5\!\times\!10^{-4}$.
We use SGD with momentum 0.9 throughout all experiments.
Unless noted, results are averaged over three independent runs.
For AGS, we set $g_{\min}=0.75$, $g_{\max}=1.10$, $c=0.5$, and $\tau=0.15$. We use a knowledge-distillation temperature of $T=2$.
We show task-agnostic final accuracy (TAg), old- and new-class accuracy, and accuracy on major, medium, and minor frequency groups.
For CIFAR-100-LT and ImageNet-100-LT, the major, medium, and minor groups contain 34, 33, and 33 classes, respectively.
For Food-101-LT, the corresponding groups contain 34, 34, and 33 classes.

\begin{figure}[t]
    \centering
    \begin{subfigure}{0.48\linewidth}
        \includegraphics[width=\linewidth]{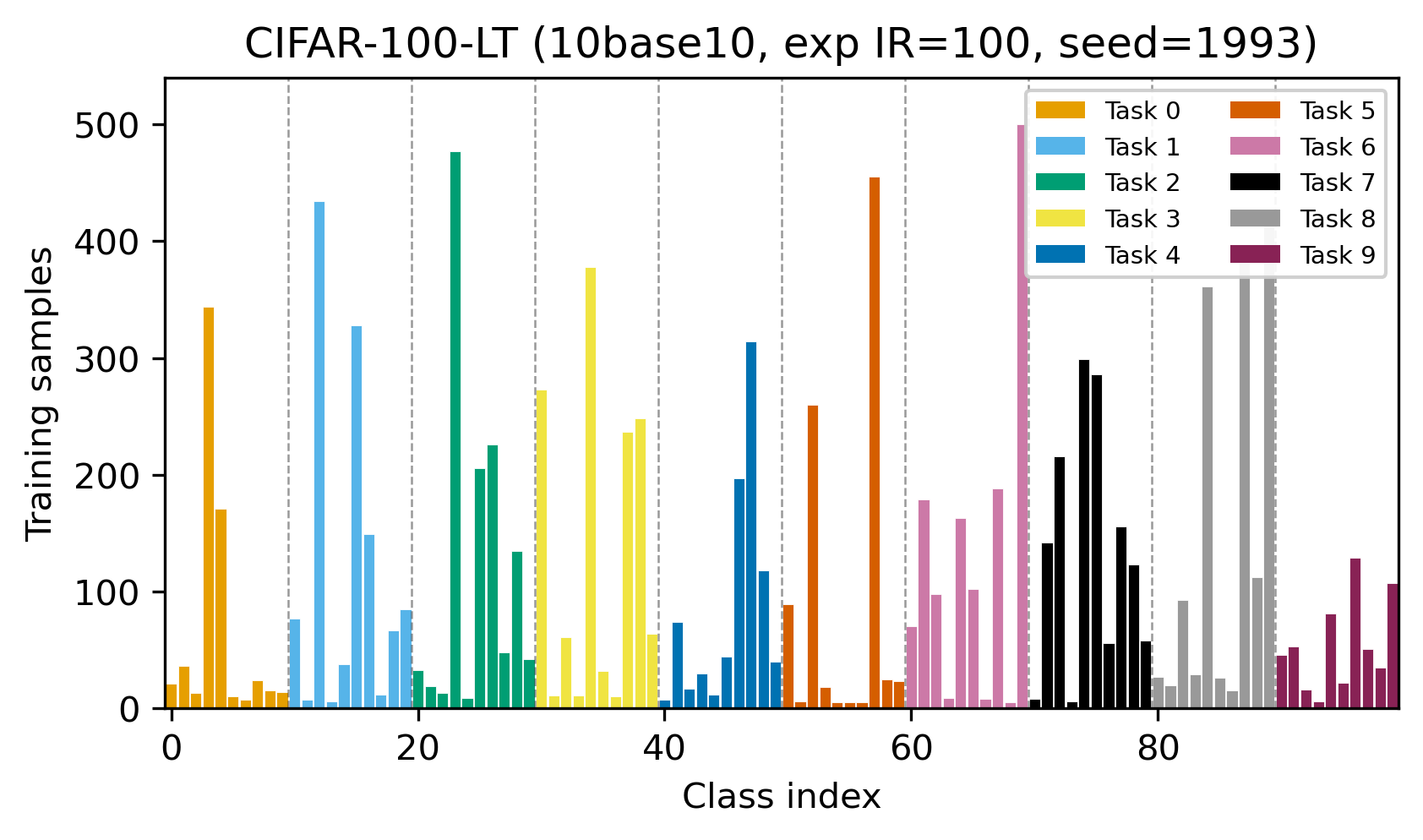}
        \caption{CIFAR-100-LT \texttt{10base10}}
    \end{subfigure}\hfill
    \begin{subfigure}{0.48\linewidth}
        \includegraphics[width=\linewidth]{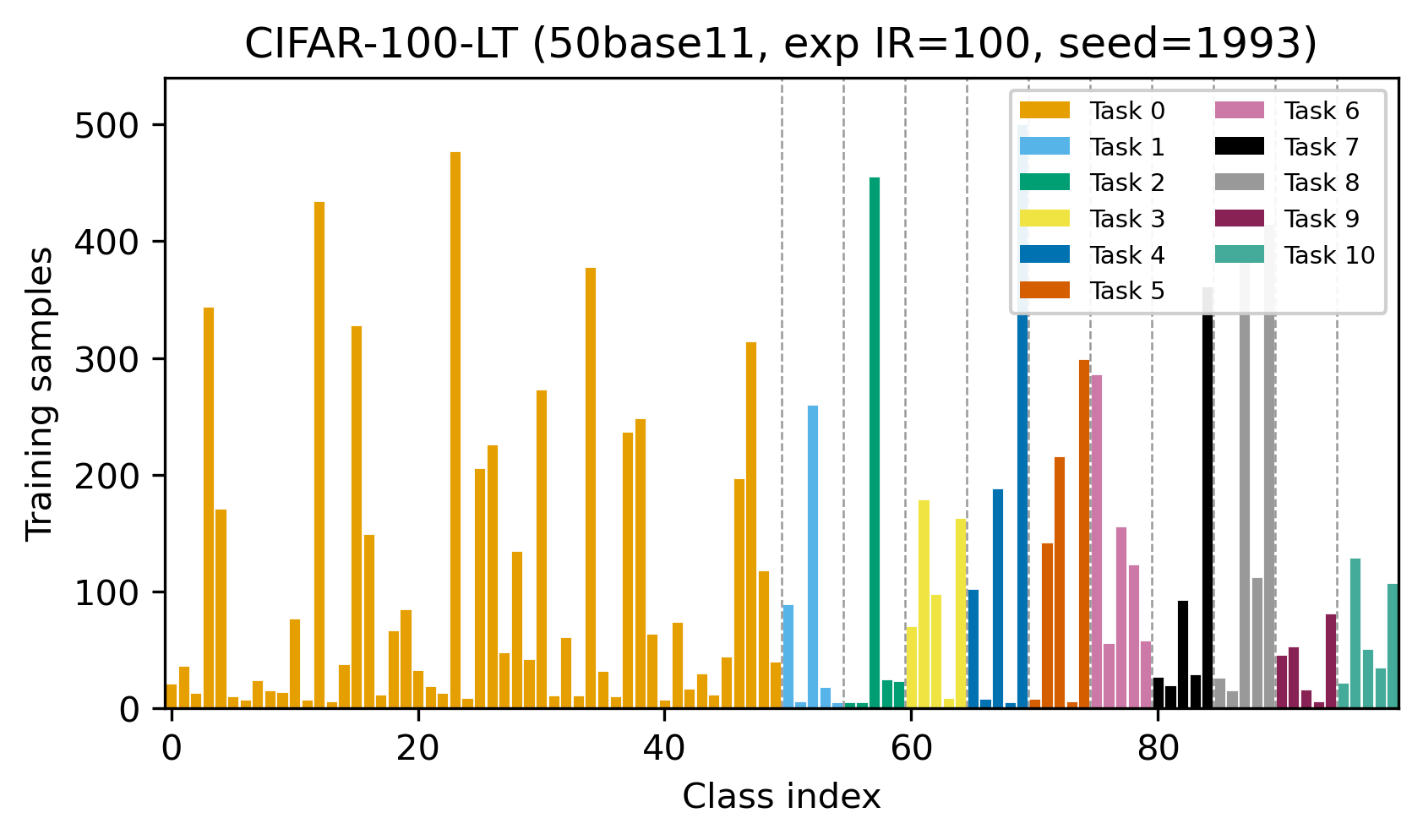}
        \caption{CIFAR-100-LT \texttt{50base11}}
    \end{subfigure}
    \\[4pt]
    \begin{subfigure}{0.48\linewidth}
        \includegraphics[width=\linewidth]{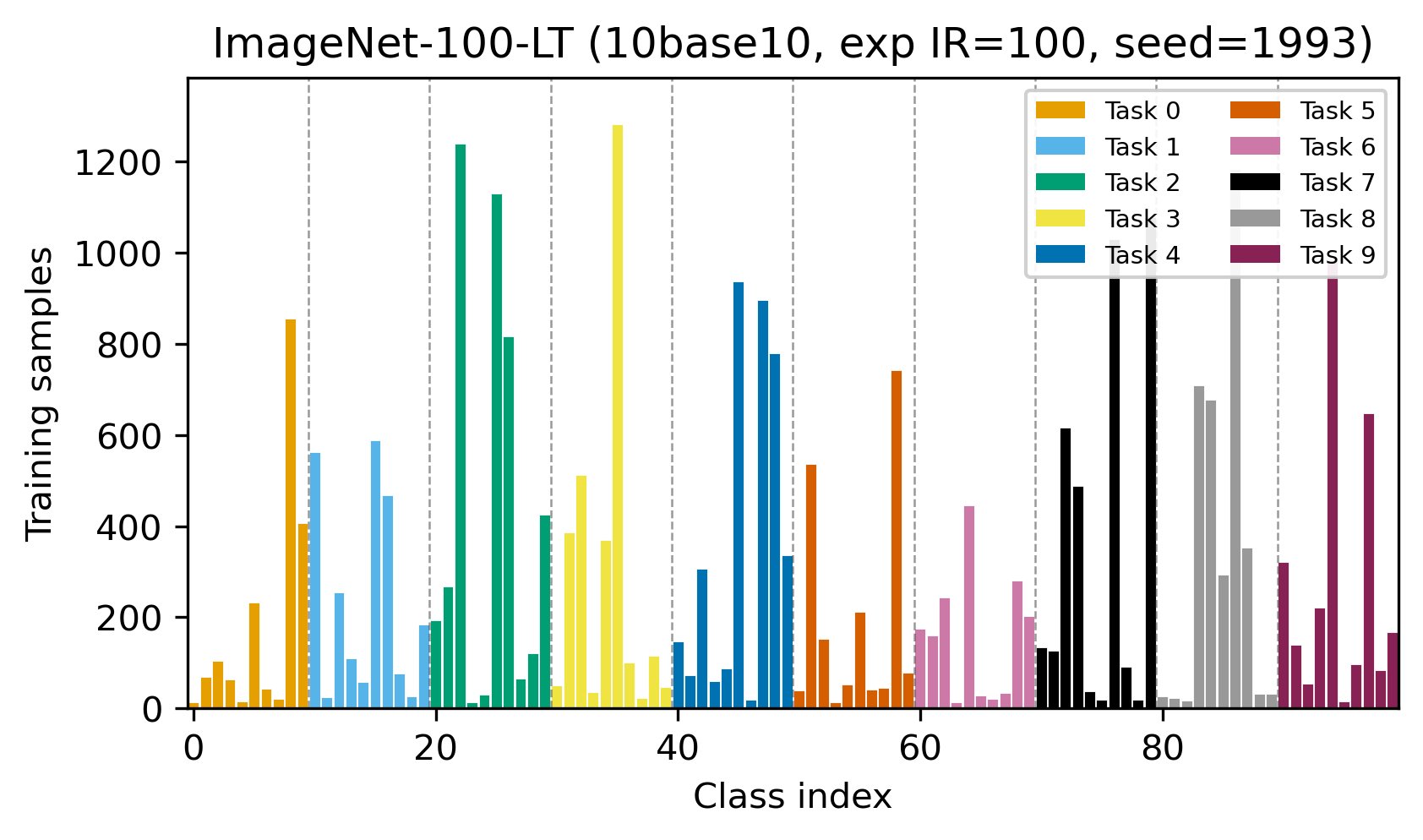}
        \caption{ImageNet-100-LT \texttt{10base10}}
    \end{subfigure}\hfill
    \begin{subfigure}{0.48\linewidth}
        \includegraphics[width=\linewidth]{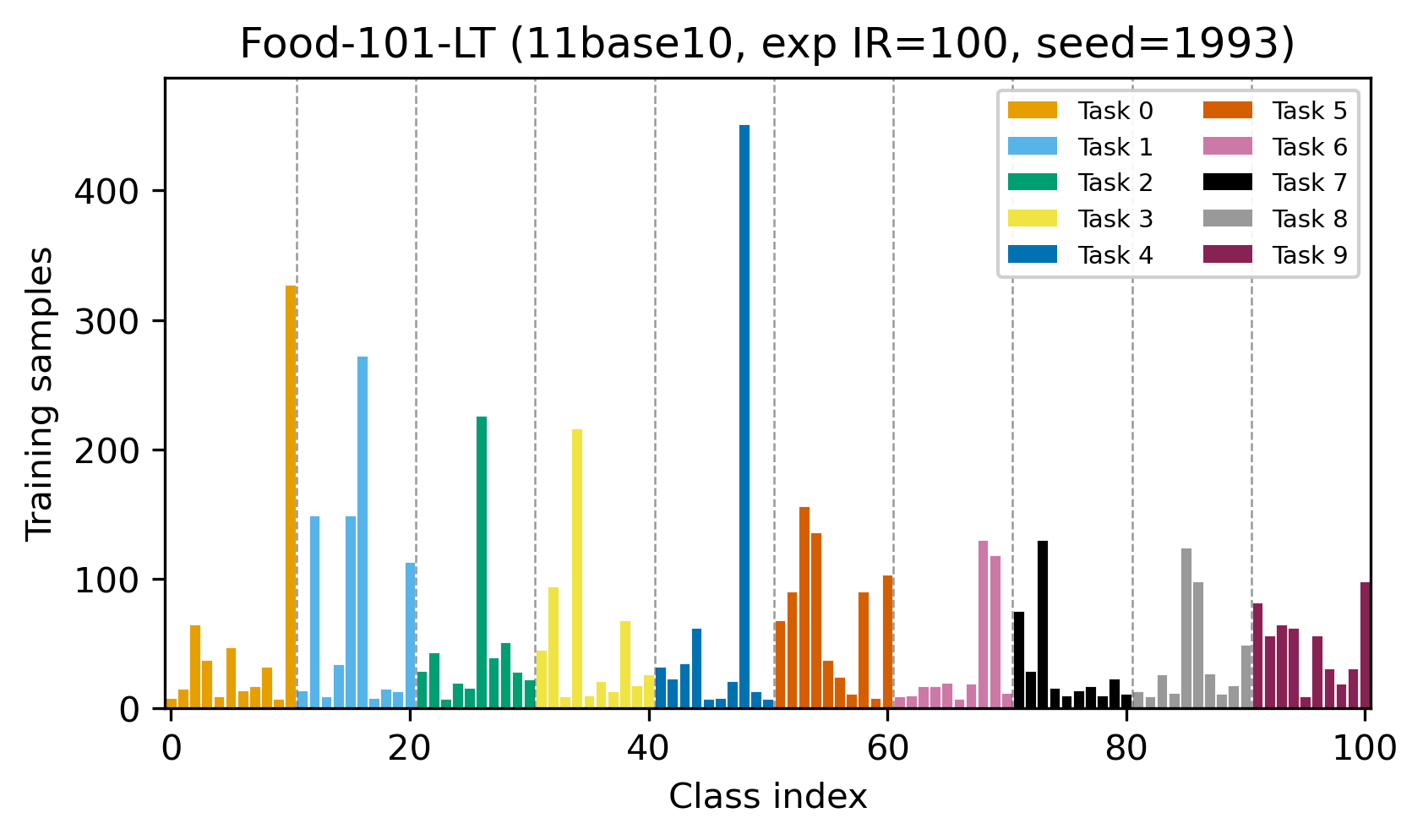}
        \caption{Food-101-LT \texttt{11base10}}
    \end{subfigure}
    \caption{Per-class training-sample distributions of the four long-tailed benchmarks. All datasets follow an exponential profile with imbalance ratio 100 (factor $0.01$), so the head classes retain the full sample count while the tail is reduced to roughly $1\%$ of it. Vertical dashed lines mark the major / medium / minor group boundaries (34\,/\,33\,/\,33 classes for CIFAR-100-LT and ImageNet-100-LT, and 34\,/\,34\,/\,33 for Food-101-LT). Class order follows the incremental task splits, so the imbalance is present both within the base task and across later increments.}
    \label{figltdist}
\end{figure}

Figure~\ref{figltdist} visualizes the resulting class-frequency profiles for the four evaluation benchmarks.
Each dataset is resampled with an exponential long-tailed profile at imbalance ratio~100, so that the most frequent class keeps its full number of training images while the rarest class retains only about $1\%$ of that count.
The steep decay is evident across all four settings. A small set of head classes dominates the sample budget, and the long tail of minor classes provides only a handful of examples each.
Because the class order is inherited from the incremental task splits, this imbalance is not confined to a single task. It recurs inside the base task and across later increments, where majority classes dominate gradient statistics and later updates can overwrite representations of rare, previously learned classes.

\subsection{Main Results}

Tables~\ref{tabcifar10}--\ref{tabfood101} compare \ours{} against DGR, DER, FOSTER, and COIL under identical replay and training schedules.
The best result in each column is shown in bold, and the second best is underlined.
On ImageNet-100-LT, our full model gives the best TAg. It improves over DGR by $0.99$ \% and over DER by $12.26$ \%, with gains in old-, major-, and medium-class accuracy.
On CIFAR-100-LT \texttt{10base10}, it reaches the best TAg, $2.35$ \% above DER. DER, an expansion-based method, remains stronger on new- and minor-class accuracy.
On \texttt{50base11}, it improves over DGR by $2.35$ \% in TAg and exceeds DER (16.96\%) by $0.32$ \%, while also outperforming FOSTER (14.05\%) and COIL (14.65\%).
On Food-101-LT, the full model gives the best TAg, improving over DGR by $1.09$ \%, with gains across old-, new-, medium-, and minor-class accuracy.

\begin{table}[t]
    \centering
    \renewcommand{\arraystretch}{1.12}
    \caption{
    Final accuracy (\%) on CIFAR-100-LT \texttt{10base10}.
        Results are averaged over three runs.
        The best result in each column is shown in bold and the second best is underlined.
    }
    \label{tabcifar10}
            \begin{tabular}{l|rrrrrr}
            \toprule
            \textbf{Method} & \textbf{TAg} & \textbf{Old} & \textbf{New} & \textbf{Major} & \textbf{Medium} & \textbf{Minor} \\
            \midrule
            COIL   & 14.06 &  9.30 & \underline{56.90} &  3.32 &  6.70 & 32.48 \\
            DER    & \underline{21.80} & 17.26 & \textbf{62.63} &  8.30 & 15.73 & \textbf{41.79} \\
            FOSTER & 12.57 & 13.95 &  0.13 & \textbf{18.03} & 14.47 &  5.05 \\
            DGR    & 21.62 & \underline{18.62} & 48.67 & 12.12 & \underline{18.39} & 34.65 \\
            NeuroGuard & \textbf{24.15} & \textbf{20.86} & 53.71 & \underline{12.94} & \textbf{21.80} & \underline{38.03} \\
            \bottomrule
        \end{tabular}
\end{table}

\begin{table}[t]
    \centering
    \renewcommand{\arraystretch}{1.12}
    \caption{
    Final accuracy (\%) on CIFAR-100-LT \texttt{50base11}.
        Results are averaged over three runs.
        The best result in each column is shown in bold and the second best is underlined.
    }
    \label{tabcifar50}
            \begin{tabular}{l|rrrrrr}
            \toprule
            \textbf{Method} & \textbf{TAg} & \textbf{Old} & \textbf{New} & \textbf{Major} & \textbf{Medium} & \textbf{Minor} \\
            \midrule
            COIL   & 14.65 & 11.91 & \textbf{66.80} &  6.39 &  8.13 & \textbf{29.68} \\
            DER    & \underline{16.96} & 14.40 & \underline{47.60} & 12.04 & 12.59 & \underline{25.46} \\
            FOSTER & 14.05 & \underline{15.25} &  0.05 &   \textbf{15.44} & \underline{15.65} & 13.04 \\
            DGR    & 14.93 & 13.80 & 36.27 & 11.69 & 12.14 & 21.05 \\
            NeuroGuard(Ours) & \textbf{17.28} & \textbf{16.02} & 41.33 & \underline{12.33} & \textbf{15.88} & 23.79 \\
            \bottomrule
        \end{tabular}
\end{table}

\begin{table}[t]
    \centering
    \renewcommand{\arraystretch}{1.12}
    \caption{
    Final accuracy (\%) on ImageNet-100-LT \texttt{10base10}.
        Results are averaged over three runs.
        The best result in each column is shown in bold and the second best is underlined.
    }
    \label{tabimagenet}
            \begin{tabular}{l|rrrrrr}
            \toprule
            \textbf{Method} & \textbf{TAg} & \textbf{Old} & \textbf{New} & \textbf{Major} & \textbf{Medium} & \textbf{Minor} \\
            \midrule
            COIL   & 17.21 & 12.16 & \underline{62.73} &  9.90 & 10.47 & 31.49 \\
            DER    & 21.77 & 16.65 & \textbf{67.80} & 12.41 & 18.06 & 35.11 \\
            FOSTER &  5.27 &  5.73 &  1.20 & 12.33 &  2.81 &  0.46 \\
            DGR    & \underline{33.04} & \underline{30.76} & 53.60 & \underline{26.51} & \underline{30.10} & \underline{42.71} \\
            NeuroGuard(Ours) & \textbf{34.03} & \textbf{31.85} & 53.73 & \textbf{27.41} & \textbf{32.14} & \textbf{42.75} \\
            \bottomrule
        \end{tabular}
\end{table}

\begin{table}[t]
    \centering
    \renewcommand{\arraystretch}{1.12}
    \caption{
    Final accuracy (\%) on Food-101-LT \texttt{11base10}.
        Results are averaged over three runs.
        The best result in each column is shown in bold and the second best is underlined.
    }
    \label{tabfood101}
            \begin{tabular}{l|rrrrrr}
            \toprule
            \textbf{Method} & \textbf{TAg} & \textbf{Old} & \textbf{New} & \textbf{Major} & \textbf{Medium} & \textbf{Minor} \\
            \midrule
            COIL   &  4.90 &  3.77 & 15.24 &  3.74 &  4.63 &  6.24 \\
            DER    &  7.86 &  4.23 & \textbf{40.85} &  3.77 &  3.28 & 15.60 \\
            FOSTER &  2.59 &  2.41 &  4.16 &  3.50 &  2.80 &  1.44 \\
            DGR    & \underline{14.05} & \underline{12.00} & 32.75 &  \textbf{9.39} & \underline{10.57} & \underline{21.92} \\
            NeuroGuard & \textbf{15.14} & \textbf{12.88} & \underline{35.61} & \textbf{9.39} & \textbf{11.32} & \textbf{24.32} \\
            \bottomrule
        \end{tabular}
\end{table}

\subsection{Progressive Component Analysis and Fixed-Scale Control}

Table~\ref{tabstack} traces the progressive additions and includes the fixed-scale control ($g{=}0.864$, the mean task-wise AGS scale across the five evaluated settings). In the \texttt{50base6} case study, AGS improves every reported group over DGR, including a $2.61$ percentage-point gain in TAg. Adding CRK and FBE gives another $0.99$ percentage-point gain in TAg while raising new- and major-class accuracy. Across the five settings, AGS exceeds DGR in four settings, and the full model exceeds AGS in all five. The fixed-scale control remains below AGS in all five settings by $0.31$, $1.09$, $0.85$, $2.84$, and $0.96$ \%, for an average gap of $1.21$ points. Since AGS and the fixed-scale control differ only in whether $g$ changes across task boundaries, this consistent result supports boundary-dependent update strength over one universal amount of gradient suppression.

\begin{table}[t]
    \centering
    \renewcommand{\arraystretch}{1.12}
    \caption{
    Progressive ablation from DGR to the full model, with a fixed-scale control.
    The fixed-scale control uses a single constant, $g=0.864$, equal to the mean task-wise AGS scale across the five evaluated settings, as defined in Sec.~\ref{sec:fixedscale}.
    The best result in each column is shown in bold and the second best is underlined.
    }
    \label{tabstack}
            \begin{tabular}{l|rrrrrr}
            \toprule
            \multicolumn{7}{l}{\textbf{(a) CIFAR-100-LT \texttt{50base6} case study}}\\
            \textbf{Method} & \textbf{TAg} & \textbf{Old} & \textbf{New} & \textbf{Major} & \textbf{Medium} & \textbf{Minor} \\
            \midrule
            DGR                      & 16.12 & 15.78 & 32.83 & 12.14 & 15.10 & 21.23 \\
            $+$AGS                   & \underline{18.73} & \underline{18.42} & \underline{34.00} & 14.38 & \underline{17.68} & \textbf{24.27} \\
            \quad $+$CRK$+$FBE (full model) & \textbf{19.72} & \textbf{19.35} & \textbf{38.00} & \textbf{16.10} & \textbf{19.25} & \underline{23.92} \\
            \midrule
            Fixed-scale & 18.42 & 18.12 & 31.46 & \underline{15.00} & 17.66 & 22.54 \\
            \bottomrule
        \end{tabular}
    \\[4pt]
            \begin{tabular}{@{}l|rrrrr@{}}
            \toprule
            \multicolumn{6}{@{}l}{\textbf{(b) TAg across settings}}\\
            \multirow{2}{*}{\textbf{Method}} & \multicolumn{3}{c}{\textbf{CIFAR-100-LT}} & \shortstack{\textbf{ImageNet-}\\\textbf{100-LT}} & \shortstack{\textbf{Food-101-}\\\textbf{LT}} \\
            \cmidrule(lr){2-4}\cmidrule(lr){5-5}\cmidrule(lr){6-6}
            & \texttt{50base6} & \texttt{10base10} & \texttt{50base11} & \texttt{10base10} & \texttt{11base10} \\
            \midrule
            DGR                  & 16.12 & 21.62 & 14.93 & 33.04 & \underline{14.05} \\
            $+$AGS               & \underline{18.73} & \underline{21.80} & \underline{16.96} & \underline{33.99} & 13.56 \\
            AGS$+$CRK$+$FBE & \textbf{19.72} & \textbf{24.15} & \textbf{17.28} & \textbf{34.03} & \textbf{15.14} \\
            \midrule
            Fixed-scale & 18.42 & 20.71 & 16.11 & 31.15 & 12.60 \\
            \bottomrule
        \end{tabular}
\end{table}

\begin{figure}[t]
    \centering
    \includegraphics[width=0.66\linewidth]{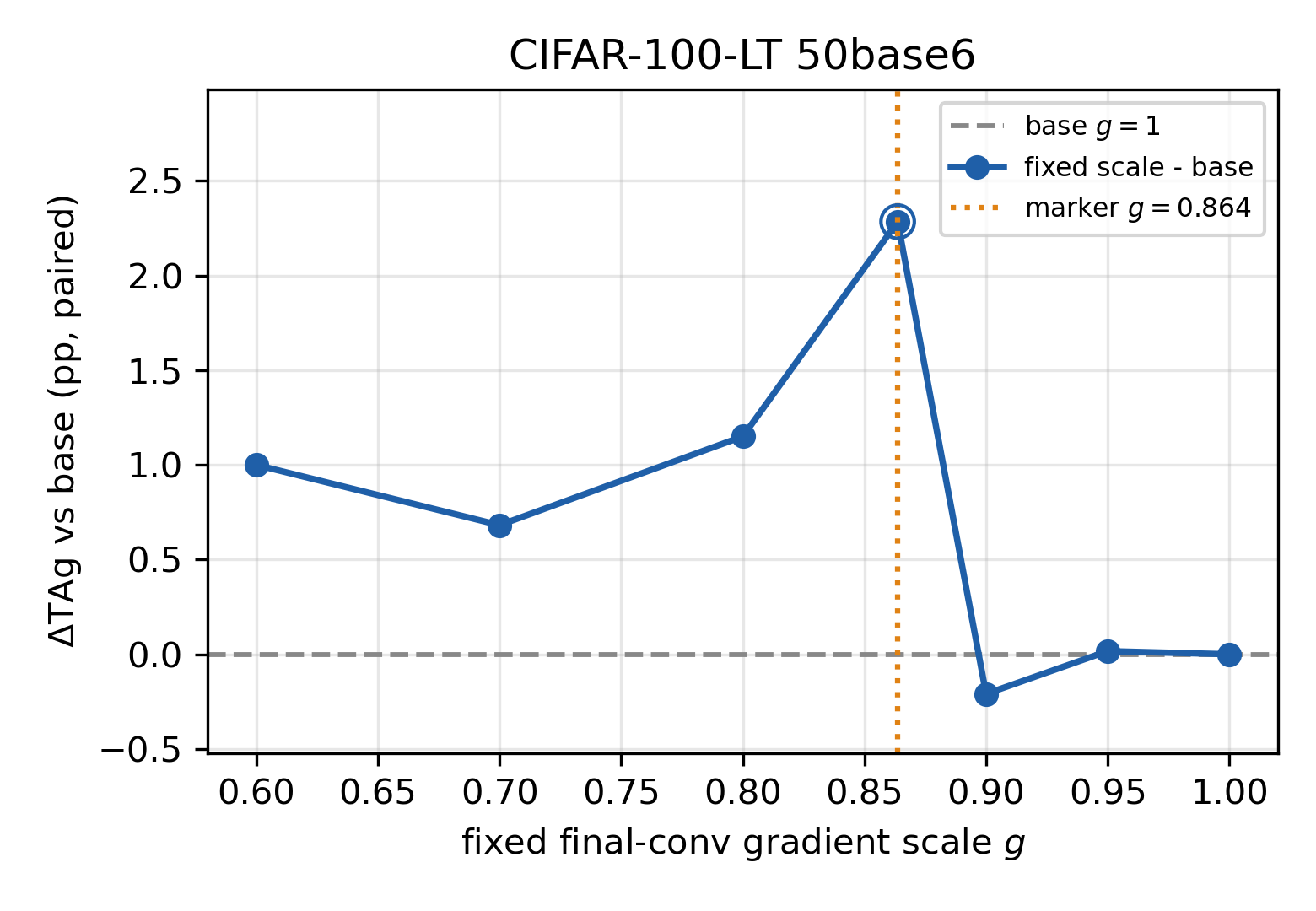}
    \caption{Fixed-scale sweep on CIFAR-100-LT \texttt{50base6} only.
    Each point is the mean TAg change from scaling the gradient of the final convolutional stage by a
    constant $g$, relative to no scaling ($g{=}1.0$) for the same run, which
    cancels run-level offsets on this split. All tested scales from $0.60$ to
    $0.864$ improve TAg, with the largest gain of $2.30$ percentage points
    at $g=0.864$ (dotted), while $g{\ge}0.90$ provides essentially no benefit.
    The same sweep on \texttt{10base10} (not shown) stays within $\pm0.7$
    percentage points in TAg for all $g$, so the useful scale range is
    setting-dependent, which motivates adaptive scale selection.}
    \label{figsweep}
\end{figure}

To test whether the benefit of AGS comes from adapting the scale at each task boundary or simply from scaling down the gradient, we sweep a constant scale $g$ on the final convolutional stage for \texttt{50base6} as shown in Fig.~\ref{figsweep}.
We compare each run with its own $g{=}1.0$ (no-scaling) baseline, which removes run-level offsets and isolates the effect of $g$ itself.
All tested scales from $0.60$ to $0.864$ improve TAg relative to their paired $g{=}1.0$ baselines, with the largest gain of $2.30$ percentage points at $g=0.864$. The tested values with $g\ge0.90$ provide essentially no gain.
On this split, a favorable constant recovers a substantial part of the AGS benefit, but the fixed-scale control in Table~\ref{tabstack} remains $0.31$ percentage points below AGS.

The same sweep on \texttt{10base10} (not plotted) shows a different behavior. No tested fixed $g$ changes TAg by more than $0.7$ percentage points ($|\Delta\mathrm{TAg}|<0.7$).
The useful range for the update scale is therefore setting-dependent. It is wide and low on \texttt{50base6}, and effectively absent on \texttt{10base10}.
A single mean scale therefore does not transfer across settings. AGS turns gradient scaling from a setting-specific hyperparameter into a task-boundary decision, making update suppression robust to changes in the class stream.

\subsection{Component Ablation for CRK and FBE}
\label{sec:ablation}

Table~\ref{tabckr} dissects knowledge-distillation reweighting and gate signals on the \texttt{50base6} case study. Values are from one run and are therefore not directly comparable with the three-run means in Table~\ref{tabstack}.
The table is organized as a progressive ablation. The first block adds rank-based reweighting and then FBE to the no-reweighting baseline. The second block evaluates the same gate with the alternative margin-proportional weighting. This comparison separates the effect of the weighting rule from the mere presence of reweighting.
Two effects are visible.
First, under entropy-only AGS, the rank form of Eq.~\eqref{eqcrk} outperforms the margin-proportional form by $3.48$ \% in TAg and also improves over the AGS baseline without reweighting.
Second, adding FBE changes TAg by $+0.07$ \% with rank-based CRK and by $-6.65$ \% with margin-proportional weighting.
Thus, FBE is compatible with the proposed rank-based CRK, yielding the best TAg in the table, whereas the margin-proportional alternative does not support the same combination.

\begin{table}[t]
    \centering
    \renewcommand{\arraystretch}{1.15}
    \caption{CRK/FBE component ablation on CIFAR-100-LT \texttt{50base6}, single run only. Each row adds one component on top of the row above within its block. Here ``rank KD'' denotes our proposed CRK, namely the rank-based weighting in Eq.~\eqref{eqcrk}. ``Margin KD'' denotes the alternative margin-proportional weighting compared only as a control.}
    \label{tabckr}
            \begin{tabular}{l|rrr}
            \toprule
            \textbf{Configuration} & \textbf{TAg} & \textbf{Old} & \textbf{New} \\
            \midrule
            AGS only (no KD reweighting)          & 24.72 & 24.07 & \textbf{56.50} \\
            \quad $+$ rank KD (CRK)                & \underline{25.14} & \underline{24.50} & \textbf{56.50} \\
            \quad\quad $+$ FBE gate (full model)    & \textbf{25.21} & \textbf{24.61} & \underline{54.50} \\
            \midrule
            AGS only $+$ margin KD (alternative)    & 21.66 & 20.16 & 34.20 \\
            \quad $+$ FBE gate (alternative)         & 15.01 & 14.44 & 43.00 \\
            \bottomrule
        \end{tabular}
\end{table}

\subsection{Behavioral Diagnostics}

Table~\ref{tabbehavior} reports the difference between the full model and DGR for each evaluation group.

\begin{table}[t]
    \centering
    \small
    \setlength{\tabcolsep}{3.2pt}
    \renewcommand{\arraystretch}{1.12}
    \caption{Difference between the full model and DGR in \%, averaged over runs. Positive values indicate higher accuracy for the full model.}
    \label{tabbehavior}
    \begin{tabular}{l|rrrrrr}
        \toprule
        \textbf{Setting} & \textbf{TAg} & \textbf{Old} & \textbf{New} & \textbf{Major} & \textbf{Medium} & \textbf{Minor} \\
        \midrule
        \texttt{50base6}   & $+3.60$ & $+3.57$ & $+5.17$ & $+3.96$ & $+4.15$ & $+2.69$ \\
        \texttt{10base10}  & $+2.53$ & $+2.24$ & $+5.04$ & $+0.82$ & $+3.41$ & $+3.38$ \\
        \texttt{50base11}  & $+2.35$ & $+2.22$ & $+5.06$ & $+0.64$ & $+3.74$ & $+2.74$ \\
        Food-101-LT        & $+1.09$ & $+0.88$ & $+2.86$ & $0.00$ & $+0.75$ & $+2.40$ \\
        ImageNet-100-LT    & $+0.99$ & $+1.09$ & $+0.13$ & $+0.90$ & $+2.04$ & $+0.04$ \\
        \bottomrule
    \end{tabular}
\end{table}

The full model improves TAg, old-, new-, medium-, and minor-class accuracy in all five settings. Major-class accuracy also improves or remains unchanged, showing that the gain is not confined to old-class retention.

\begin{figure}[t]
    \centering
    \includegraphics[width=\linewidth]{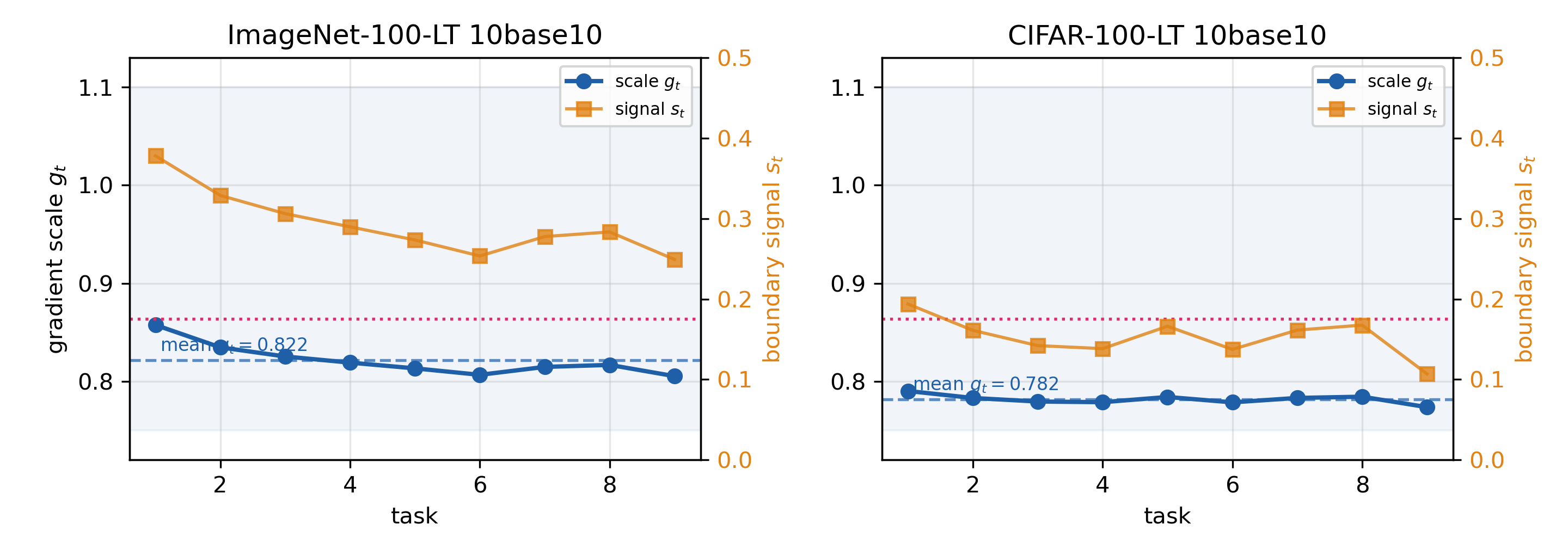}
    \caption{Realized adaptive scale $g_t$ (blue) and boundary signal $s_t$ (orange) per task for NeuroGuard. The scale tracks the signal and stays inside the AGS range (shaded), near the mean task-wise AGS scale across the five evaluated settings ($g{=}0.864$, dotted). It varies across tasks on ImageNet-100-LT (mean $0.82$) and remains nearly constant on CIFAR-100-LT \texttt{10base10} (mean $0.78$), consistent with the flat fixed-scale sweep on that split.}
    \label{figtraj}
\end{figure}

Figure~\ref{figtraj} shows that the realized scale $g_t$ tracks the boundary signal instead of collapsing to a constant.
The scale adapts across tasks on ImageNet-100-LT and stays nearly flat on \texttt{10base10}, consistent with the flat sweep on that split.

\section{Limitations}
\label{seclimitations}

Our update control follows the neuromodulation-inspired motivation described in Sec.~\ref{secintro}, but the current formulation represents uncertainty-dependent plasticity with one task-boundary signal and one gradient scale. Richer temporal and layer-wise regulation remains future work. The evaluation also assumes known task boundaries and a fixed replay budget of five exemplars per class.

\section{Conclusion}

We presented NeuroGuard, a fragility-aware update-control strategy for long-tailed class-incremental learning.
Adaptive Gradient Scaling (AGS) controls representation updates using task-boundary uncertainty.
Confidence-Ranked Knowledge Distillation Reweighting (CRK) assigns larger distillation weights to replayed memories that the teacher considers fragile while keeping the mean weight equal to one.
The Fragility-Blended Entropy Gate (FBE) augments the boundary signal with a simple measure of old-memory leakage.
Together, these components introduce no additional learnable parameters, do not add a loss term, and leave the replay memory and classifier unchanged.

Across five LT-CIL settings, the method consistently improves over the DGR baseline and achieves the best task-agnostic accuracy among all compared methods on all four benchmark comparisons.
The fixed-scale control using $g=0.864$ remains below AGS in all five settings by an average of $1.21$ \% and below DGR in three, even though its value is computed from the AGS runs themselves. The \texttt{50base6} sweep shows that a favorable constant can recover part of the gain on one split, whereas the same constant does not transfer across settings. Matching the mean amount of suppression is therefore insufficient; adapting update strength to each task boundary provides the consistent advantage.
Future work will investigate principled scale-selection strategies, richer boundary signals, and layer-wise update control.

\bibliographystyle{splncs04}
\bibliography{main}

@String(CVPR  = {IEEE Conf. Comput. Vis. Pattern Recog.})

@String(ECCV  = {Eur. Conf. Comput. Vis.})

@String(ICML  = {Int. Conf. Mach. Learn.})

@String(CVPR  = {CVPR})

@String(ECCV  = {ECCV})

@String(ICML  = {ICML})

@InProceedings{liu2022longtailed,
author = {Liu, Xialei and Hu, Yu-Song and Cao, Xu-Sheng and Bagdanov, Andrew D. and Li, Ke and Cheng, Ming-Ming},
editor="Avidan, Shai
and Brostow, Gabriel
and Ciss{\'e}, Moustapha
and Farinella, Giovanni Maria
and Hassner, Tal",
title="Long-Tailed Class Incremental Learning",
booktitle="Computer Vision -- ECCV 2022",
year="2022",
pages="495--512",
isbn="978-3-031-19827-4"
}

@inproceedings{menon2021longtail,
title	= {Long-tail learning via logit adjustment},
author = {Menon, Aditya Krishna and Veit, Andreas and Rawat, Ankit Singh and Jain, Himanshu and Jayasumana, Sadeep and Kumar, Sanjiv},
year	= {2021},
booktitle	= {International Conference on Learning Representations}}

@inproceedings{kang2021balancedkd,
  title     = {{Exploring Balanced Feature Spaces for Representation Learning}},
  author = {Kang, Bingyi and Li, Yu and Xie, Sa and Yuan, Zehuan and Feng, Jiashi},
  booktitle = {International Conference on Learning Representations},
  year      = {2021},
}

@inproceedings{suh2023ltmi,
  title     = {Long-Tailed Recognition by Mutual Information Maximization between Latent Features and Ground-Truth Labels},
  author = {Suh, Yubin and Kim, Hyunsoo and Han, Bohyung},
  booktitle = {ICML},
  year      = {2023},
}

@INPROCEEDINGS{mdcs2023,
  author = {Zhao, Qihao and Jiang, Chen and Hu, Wei and Zhang, Fan and Liu, Jun},
  booktitle={IEEE/CVF International Conference on Computer Vision}, 
  title={MDCS: More Diverse Experts with Consistency Self-distillation for Long-tailed Recognition}, 
  year={2023},
  volume={},
  number={},
  pages={11563-11574},
  doi={10.1109/ICCV51070.2023.01065}}

@article{ltfood2023,
  title={Long-tailed continual learning for visual food recognition},
  author = {He, Jiangpeng and Zhang, Xiaoyan and Lin, Luotao and Ma, Jack and Eicher-Miller, Heather A and Zhu, Fengqing},
  journal={IEEE transactions on multimedia},
  volume={28},
  pages={865--877},
  year={2025},
  publisher={IEEE}
}

@article{opencil2024,
author = {Miao, Wenjun and Pang, Guansong and Nguyen, Trong-Tung and Fang, Ruohuan and Zheng, Jin and Bai, Xiao},
title = {OpenCIL: Benchmarking out-of-distribution detection in class incremental learning},
year = {2026},
volume = {171},
number = {PA},
issn = {0031-3203},
doi = {10.1016/j.patcog.2025.112163},
journal = {Pattern Recogn.},
numpages = {17},
}

@ARTICLE{masana2023survey,
  author = {Zhou, Da-Wei and Wang, Qi-Wei and Qi, Zhi-Hong and Ye, Han-Jia and Zhan, De-Chuan and Liu, Ziwei},
  journal={IEEE Transactions on Pattern Analysis and Machine Intelligence}, 
  title={Class-Incremental Learning: A Survey}, 
  year={2024},
  volume={46},
  number={12},
  pages={9851-9873},
  doi={10.1109/TPAMI.2024.3429383}
  }

@article{vandeven2022three,
  author = {van de Ven, Gido M. and Tuytelaars, Tinne and Tolias, Andreas S.},
  title = {Three types of incremental learning},
  journal = {Nature Machine Intelligence},
  year = {2022},
  volume = {4},
  number = {12},
  pages = {1185--1197},
  month = dec,
  doi = {10.1038/s42256-022-00568-3},
  issn = {2522-5839}
}

@ARTICLE{delange2021isCILenough,
  author = {De Lange, Matthias and Aljundi, Rahaf and Masana, Marc and Parisot, Sarah and Jia, Xu and Leonardis, Aleš and Slabaugh, Gregory and Tuytelaars, Tinne},
  journal={IEEE Transactions on Pattern Analysis and Machine Intelligence}, 
  title={A Continual Learning Survey: Defying Forgetting in Classification Tasks}, 
  year={2022},
  volume={44},
  number={7},
  pages={3366-3385},
  doi={10.1109/TPAMI.2021.3057446}
  }

@article{mundt2021essentials,
title = {A wholistic view of continual learning with deep neural networks: Forgotten lessons and the bridge to active and open world learning},
journal = {Neural Networks},
volume = {160},
pages = {306-336},
year = {2023},
issn = {0893-6080},
doi = {https://doi.org/10.1016/j.neunet.2023.01.014},
author = {Mundt, Martin and Hong, Yongwon and Pliushch, Iuliia and Ramesh, Visvanathan},
}

@article{dgr,
  title={Gradient Reweighting: Towards Imbalanced Class-Incremental Learning},
  author = {He, Jiangpeng and Zhu, Fengqing Maggie},
  journal={IEEE/CVF Conference on Computer Vision and Pattern Recognition},
  year={2024},
  pages={16668-16677},
}

@inproceedings{yan2021der,
  title={Der: Dynamically expandable representation for class incremental learning},
  author = {Yan, Shipeng and Xie, Jiangwei and He, Xuming},
  booktitle={Proceedings of the IEEE/CVF conference on computer vision and pattern recognition},
  pages={3014--3023},
  year={2021}
}

@inproceedings{wang2022foster,
author = {Wang, Fu-Yun and Zhou, Da-Wei and Ye, Han-Jia and Zhan, De-Chuan},
title = {FOSTER: Feature Boosting and Compression for Class-Incremental Learning},
year = {2022},
isbn = {978-3-031-19805-2},
doi = {10.1007/978-3-031-19806-9_23},
booktitle = {Computer Vision – ECCV 2022: 17th European Conference, Tel Aviv, Israel, October 23–27, 2022, Proceedings, Part XXV},
pages = {398–414},
numpages = {17},
}

@inproceedings{coil2021,
author = {Zhou, Da-Wei and Ye, Han-Jia and Zhan, De-Chuan},
title = {Co-Transport for Class-Incremental Learning},
year = {2021},
isbn = {9781450386517},
doi = {10.1145/3474085.3475306},
booktitle = {Proceedings of the 29th ACM International Conference on Multimedia},
pages = {1645–1654},
numpages = {10},
}

@INPROCEEDINGS{rebuffi2017icarl,
  author = {Rebuffi, Sylvestre-Alvise and Kolesnikov, Alexander and Sperl, Georg and Lampert, Christoph H.},
  booktitle={IEEE Conference on Computer Vision and Pattern Recognition}, 
  title={iCaRL: Incremental Classifier and Representation Learning}, 
  year={2017},
  volume={},
  number={},
  pages={5533-5542},
  doi={10.1109/CVPR.2017.587}
}

@INPROCEEDINGS{hou2019learning,
  author = {Hou, Saihui and Pan, Xinyu and Loy, Chen Change and Wang, Zilei and Lin, Dahua},
  booktitle={IEEE/CVF Conference on Computer Vision and Pattern Recognition}, 
  title={Learning a Unified Classifier Incrementally via Rebalancing}, 
  year={2019},
  volume={},
  number={},
  pages={831-839},
  doi={10.1109/CVPR.2019.00092}
  }

@InProceedings{kim2020imbalanced,
author = {Kim, Chris Dongjoo and Jeong, Jinseo and Kim, Gunhee},
title="Imbalanced Continual Learning with Partitioning Reservoir Sampling",
booktitle="Computer Vision -- ECCV 2020",
year="2020",
pages="411--428",
isbn="978-3-030-58601-0"
}

@INPROCEEDINGS{tbbn2022,
  author = {Cha, Sungmin and Cho, Sungjun and Hwang, Dasol and Hong, Sunwon and Lee, Moontae and Moon, Taesup},
  booktitle={IEEE/CVF Conference on Computer Vision and Pattern Recognition (CVPR)}, 
  title={Rebalancing Batch Normalization for Exemplar-Based Class-Incremental Learning}, 
  year={2023},
  volume={},
  number={},
  pages={20127-20136},
  doi={10.1109/CVPR52729.2023.01927}
  }

@article{futureproof2024,
  author = {Jodelet, Quentin and Liu, Xin and Phua, Yin Jun and Murata, Tsuyoshi},
  title = {Future-proofing class-incremental learning},
  journal = {Machine Vision and Applications},
  year = {2024},
  volume = {36},
  number = {1},
  pages = {16},
  doi = {10.1007/s00138-024-01635-y},
  issn = {1432-1769},
}

@InProceedings{primeaware2020,
author = {Zhang, Youcai and Lan, Zhonghao and Dai, Yuchen and Zeng, Fangao and Bai, Yan and Chang, Jie and Wei, Yichen},
title="Prime-Aware Adaptive Distillation",
booktitle="Computer Vision -- ECCV 2020",
year="2020",
publisher="Springer International Publishing",
pages="658--674",
isbn="978-3-030-58529-7"
}

@article{subspaceKD2023,
title = {Subspace distillation for continual learning},
journal = {Neural Networks},
volume = {167},
pages = {65-79},
year = {2023},
issn = {0893-6080},
doi = {https://doi.org/10.1016/j.neunet.2023.07.047},
author = {Roy, Kaushik and Simon, Christian and Moghadam, Peyman and Harandi, Mehrtash},
}

@ARTICLE{clkdsurvey2023,
  author = {Li, Songze and Su, Tonghua and Zhang, Xu-Yao and Wang, Zhongjie},
  journal={IEEE Transactions on Neural Networks and Learning Systems}, 
  title={Continual Learning With Knowledge Distillation: A Survey}, 
  year={2025},
  volume={36},
  number={6},
  pages={9798-9818},
  doi={10.1109/TNNLS.2024.3476068}
  }

@InProceedings{pgd_crs2021,
author = {Chen, Zhiyi and Lin, Tong},
title="Principal Gradient Direction and Confidence Reservoir Sampling for Continual Learning",
booktitle="Artificial Neural Networks and Machine Learning -- ICANN 2021",
year="2021",
publisher="Springer International Publishing",
pages="421--432",
isbn="978-3-030-86340-1"
}

@inproceedings{er2019,
 author = {Rolnick, David and Ahuja, Arun and Schwarz, Jonathan and Lillicrap, Timothy and Wayne, Gregory},
 booktitle = {Advances in Neural Information Processing Systems},
 pages = {},
 title = {Experience Replay for Continual Learning},
 volume = {32},
 year = {2019}
}

@article{kirkpatrick2017overcoming,
  title={Overcoming catastrophic forgetting in neural networks},
  author = {Kirkpatrick, James and Pascanu, Razvan and Rabinowitz, Neil and Veness, Joel and Desjardins, Guillaume and Rusu, Andrei A and Milan, Kieran and Quan, John and Ramalho, Tiago and Grabska-Barwinska, Agnieszka and others},
  journal={Proceedings of the national academy of sciences},
  volume={114},
  number={13},
  pages={3521--3526},
  year={2017},
}

@InProceedings{zenke2017continual,
  title = 	 {Continual Learning Through Synaptic Intelligence},
  author = {Zenke, Friedemann and Poole, Ben and Ganguli, Surya},
  booktitle = 	 {Proceedings of the 34th International Conference on Machine Learning},
  pages = 	 {3987--3995},
  year = 	 {2017},
  volume = 	 {70},
  month = 	 {06--11 Aug},
}

@article{laborieux2021synaptic,
  author = {Laborieux, Axel and Ernoult, Maxence and Hirtzlin, Tifenn and Querlioz, Damien},
  title = {Synaptic metaplasticity in binarized neural networks},
  journal = {Nature Communications},
  year = {2021},
  volume = {12},
  number = {1},
  pages = {2549},
  doi = {10.1038/s41467-021-22768-y},
}

@article{abraham1996metaplasticity,
title = {Metaplasticity: the plasticity of synaptic plasticity},
journal = {Trends in Neurosciences},
volume = {19},
number = {4},
pages = {126-130},
year = {1996},
issn = {0166-2236},
doi = {https://doi.org/10.1016/S0166-2236(96)80018-X},
author = {Abraham, Wickliffe C. and Bear, Mark F.},
}

@mastersthesis{cifar100lt,
  author = {Krizhevsky, Alex},
  title = {Learning Multiple Layers of Features from Tiny Images},
  school = {University of Toronto},
  year = {2009},
}

@InProceedings{liu2022code,
author = {Liu, Xialei and Hu, Yu-Song and Cao, Xu-Sheng and Bagdanov, Andrew D. and Li, Ke and Cheng, Ming-Ming},
title="Long-Tailed Class Incremental Learning",
booktitle="Computer Vision -- ECCV 2022",
year="2022",
pages="495--512",
isbn="978-3-031-19827-4"
}

@article{clapps2023,
  title={Continual Learning: Applications and the Road Forward},
  author = {Verwimp, E and Aljundi, R and Ben-David, S and Bethge, M and Cossu, A and Gepperth, A and Hayes, TL and Hullermeier, E and Kanan, C and Kudithipudi, D and others},
  journal={TRANSACTIONS ON MACHINE LEARNING RESEARCH},
  volume={2024},
  year={2024}
}

@inproceedings{saha2021gradient,
  title     = {Gradient Projection Memory for Continual Learning},
  author = {Saha, Gobinda and Garg, Isha and Roy, Kausik},
  booktitle = {9th International Conference on Learning Representations Conference Track Proceedings},
  year      = {2021}
}

@article{lopezpaz2017gem,
  title={Gradient episodic memory for continual learning},
  author = {Lopez-Paz, David and Ranzato, Marc'Aurelio},
  journal={Advances in neural information processing systems},
  volume={30},
  year={2017}
}

@article{astonjonescohen2005,
  author = {Aston-Jones, Gary and Cohen, Jonathan D.},
  title   = {An integrative theory of locus coeruleus-norepinephrine function: Adaptive gain and optimal performance},
  journal = {Annual Review of Neuroscience},
  volume  = {28},
  pages   = {403--450},
  year    = {2005},
  doi     = {10.1146/annurev.neuro.28.061604.135709}
}

@article{yudayan2005,
  author = {Yu, Angela J. and Dayan, Peter},
  title   = {Uncertainty, neuromodulation, and attention},
  journal = {Neuron},
  volume  = {46},
  number  = {4},
  pages   = {681--692},
  year    = {2005},
  doi     = {10.1016/j.neuron.2005.04.026}
}

\end{document}